\documentclass[letterpaper]{article} 
\usepackage{aaai24}  
\usepackage{times}  
\usepackage{helvet}  
\usepackage{courier}  
\usepackage[hyphens]{url}  
\usepackage{graphicx} 
\usepackage{natbib}  
\usepackage{caption} 
\usepackage{times}
\usepackage{epsfig}
\usepackage{graphicx}
\usepackage{amsmath}
\usepackage{amssymb}
\usepackage{multirow}
\usepackage{color}
\usepackage{booktabs}
\usepackage{caption}
\usepackage{bm}
\usepackage{subfigure}  
\usepackage{float}  
\usepackage{bbding}
\usepackage[ruled,vlined]{algorithm2e}
\usepackage{subfigure}  
\usepackage{float}  

\usepackage{newfloat}
\usepackage{listings}
\DeclareCaptionStyle{ruled}{labelfont=normalfont,labelsep=colon,strut=off} 
\floatstyle{ruled}
\newfloat{listing}{tb}{lst}{}
\floatname{listing}{Listing}
\title{Context-Aware Iteration Policy Network for Efficient Optical Flow Estimation}
\author {
    Ri Cheng,
    Ruian He,
    Xuhao Jiang,
    Shili Zhou,
    Weimin Tan\footnote{Corresponding authors: Weimin Tan, Bo Yan. This work is supported by NSFC (GrantNo.: U2001209 and 62372117) and Natural Science Foundation of Shanghai (21ZR1406600).},
    Bo Yan\footnotemark[1]
}
\affiliations {
    School of Computer Science, Shanghai Key Laboratory of Intelligent Information Processing, Fudan University\\
    rcheng22@m.fudan.edu.cn, rahe16@fudan.edu.cn, 20110240011@fudan.edu.cn, slzhou19@fudan.edu.cn, wmtan@fudan.edu.cn, byan@fudan.edu.cn
}

\usepackage{bibentry}

\begin{document}

\maketitle

\begin{abstract}



Existing recurrent optical flow estimation networks are computationally expensive since they use a fixed large number of iterations to update the flow field for each sample. An efficient network should skip iterations when the flow improvement is limited. In this paper, we develop a Context-Aware Iteration Policy Network for efficient optical flow estimation, which determines the optimal number of iterations per sample. The policy network achieves this by learning contextual information to realize whether flow improvement is bottlenecked or minimal. On the one hand, we use iteration embedding and historical hidden cell, which include previous iterations information, to convey how flow has changed from previous iterations. On the other hand, we use the incremental loss to make the policy network implicitly perceive the magnitude of optical flow improvement in the subsequent iteration. Furthermore, the computational complexity in our dynamic network is controllable, allowing us to satisfy various resource preferences with a single trained model. Our policy network can be easily integrated into state-of-the-art optical flow networks. Extensive experiments show that our method maintains performance while reducing FLOPs by about 40\%/20\% for the Sintel/KITTI datasets.


\end{abstract}

\section{Introduction}

    
    \begin{figure}[t]
    \centering
    \includegraphics[width=\linewidth]{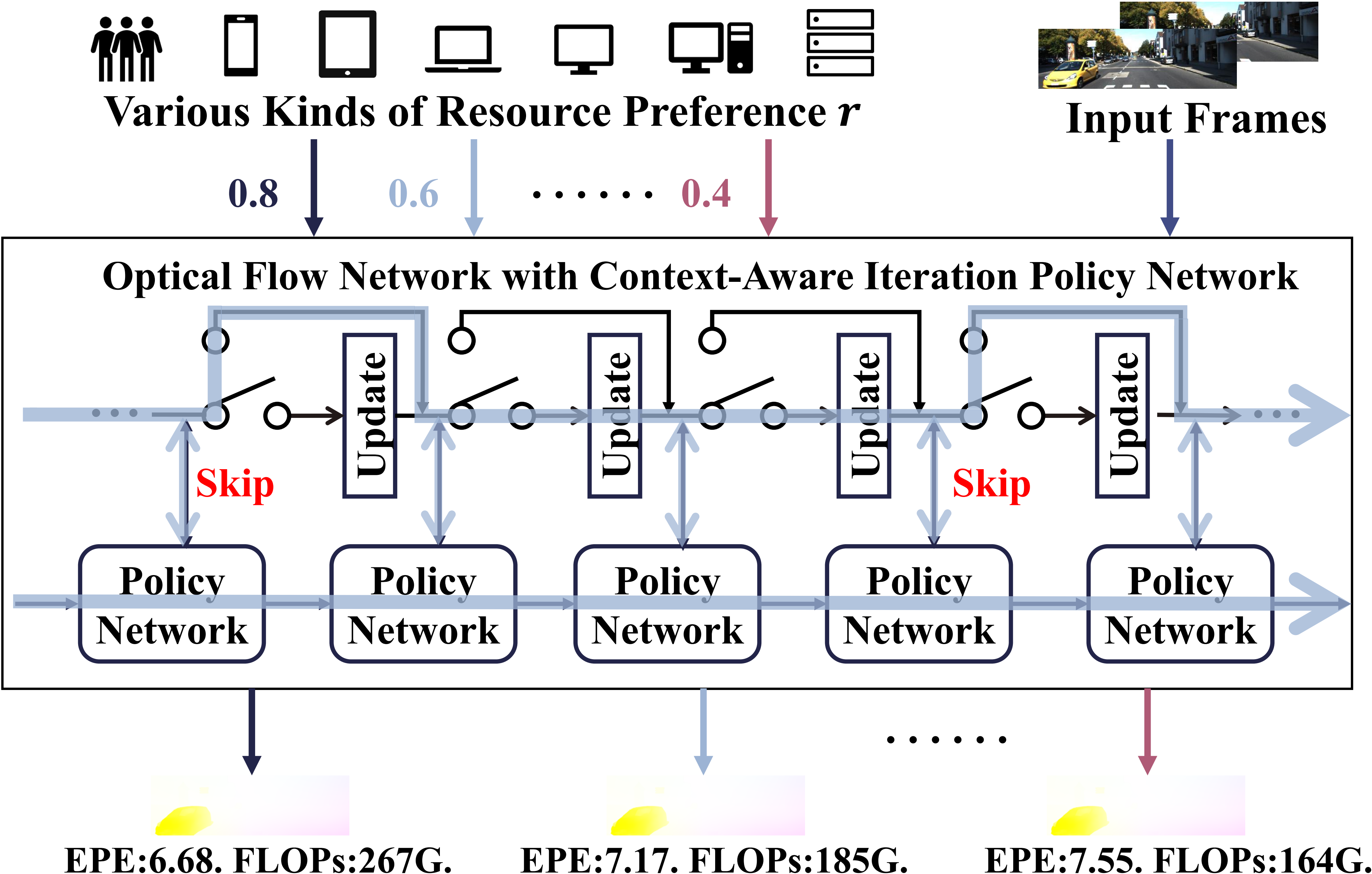}
    \caption{\textbf{Efficient inference.} Our policy network can skip the iteration to determine the optimal number of iterations depending on contextual information. Users are capable of altering the resource preference value $r$ to control computational complexity. 
    }
    \label{003_intro}
    \end{figure}


Optical flow is a fundamental task that attempts to estimate per-pixel correspondences between video frames. Optical flow models are widely used in applications such as video tracking \cite{Vihlman_Visala_2020}, video super-resolution \cite{Chan_2022_CVPR}, video frame interpolation \cite{Kong_2022_CVPR}, and autonomous driving \cite{9304777}. 
Recently, following RAFT \cite{10.1007/978-3-030-58536-5_24}, the recurrent networks have demonstrated superior performance, because they can optimize the optical flow in a iterative manner.
However, they estimate optical flow with a fixed large number of iterations, such as 32 iterations for the Sintel dataset \cite{10.1007/978-3-642-33783-3_44}, which is restrained by limited computational resource during inference.
Therefore, exploring efficient optical flow estimation algorithms is urgently needed for practical applications.

    \begin{figure*}[t]
    \centering
    \includegraphics[width=\linewidth]{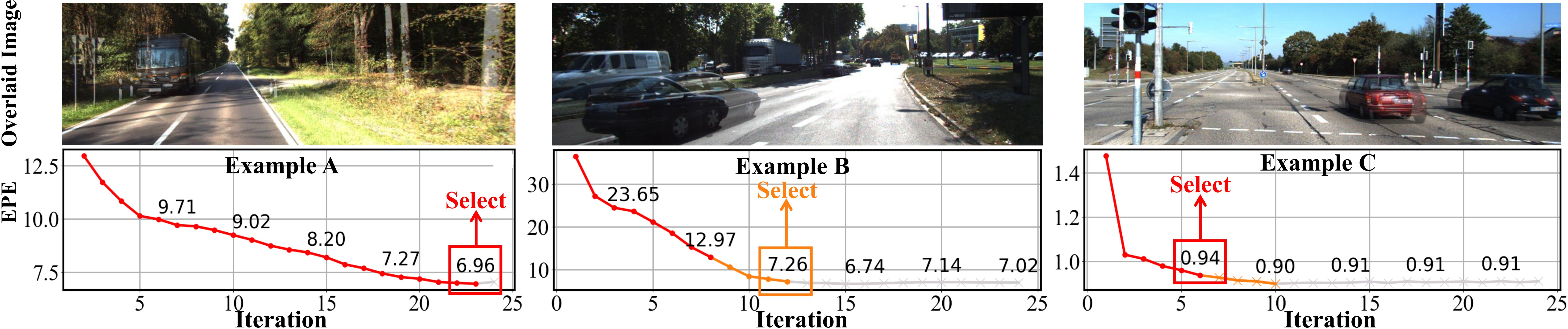}
    \caption{\textbf{Three examples of changes in EPE as iteration increases.} When the network encounters a bottleneck or the EPE improvement is small, we can reduce the computational complexity by skipping iterations. The EPE of examples B and C has a bottleneck after the 12th and 10th iterations. Compared with example A and B, the EPE improvement of example C from the 6th to 10th iteration is small.
    }
    \label{002_intro}
    \end{figure*}

To address this problem, we present three representative examples in Figure~\ref{002_intro} with two discoveries. (1)
There will be a bottleneck in the optical flow network. A bottleneck indicates that the flow network cannot improve the flow after some iterations due to the limited estimation ability of the network.
For example, the average Endpoint Error (EPE) improvement of examples B and C fails to improve EPE after the 12th and 10th iterations. 
(2) Although EPE continues to improve, the magnitude of improvement is different for each sample.
For example, example B decreases EPE by 5.71 from the 8th to 12th iteration, while example C only decreases EPE by 0.04 from the 6th to 10th iteration. Therefore, if we encounter resource constraints, we can reduce computational complexity in two ways. The first is to skip the iteration when a bottleneck is encountered, and the second is to skip an iteration where the sample only has a marginal optical flow improvement.

In this paper, we propose the dynamic optical flow network with our proposed Context-Aware Iteration Policy Network, to dynamically determine optimal number of iterations for efficient optical flow estimation. 
We integrated our proposed policy network into four state-of-the-art backbones. 
The experiment results show that our dynamic networks can maintain performance while reducing floating point operations (FLOPs) by around 20\% and 40\% for KITTI \cite{Menze_2015_CVPR} and Sintel-Final \cite{10.1007/978-3-642-33783-3_44} training datasets.
Figure~\ref{003_intro} displays the overview of our method. 
Our iteration policy determines whether to skip the iteration at each time step, and we design the policy network to be lightweight so that it does not introduce a large amount of FLOPs.
In addition, users can alter the resource preference value $r$ to control computational complexity based on the availability of various computing resources.
We force the computation cost of the recurrent process less than $r$ times the original.

However, it is challenging for the policy network to decide whether to skip iteration or not, since the features in each iteration lack the global iteration information including the process and magnitude of optical flow improvement.
Therefore, we must provide extra information to the policy network, so we propose to make the policy network rely on contextual information to decide. Specifically, historical hidden cell and iteration embedding are fed into the policy network to provide previous iterations information.
Historical hidden cell, which comes from the previous policy network, contains the information about preceding decision and flow change information, and iteration embedding provides iteration progress information. 

For future iterations information, we use the incremental loss to force the policy network to predict how much the optical flow will improve in the subsequent iteration. 
Both previous and future iterations information help the policy to realize whether the iteration network hits a bottleneck and the magnitude of optical flow improvement.
By exploiting this contextual information, the iteration policy network can determine whether to skip the iteration or not.


    This paper mainly has the following contributions:
    \begin{itemize}
    \item This paper presents an efficient optical flow estimation method that reduces computational complexity by determining the optimal number of iterations for each sample. The proposed policy network is also controllable, and it only needs a single trained network to deal with different computational resource situations.
    
    \item The proposed context-aware policy network can determine whether the flow improvement is limited. 
    We propose using historical hidden cell, iteration embedding and incremental loss to help the policy network estimate whether the flow improvement encounters the bottleneck and the magnitude of the improvement.

    \item Our policy network can be seamlessly integrated into contemporary optical flow architecture. The experiments show that our dynamic network can maintain the performance while reducing FLOPs by about 40\%/20\% for Sintel/KITTI datasets.
    \end{itemize}

\section{Related work}
\noindent
\textbf{Optical Flow Estimation.}
Optical flow aims to find pixel-wise correspondences between two video frames. 
The traditional method models optical flow as an optimization problem, and they try to improve regularizations \cite{378214,10.1007/978-3-319-10590-1_29,article1} and energy terms \cite{10.1007/978-3-540-74936-3_22} to maximize the visual similarity between image pairs.

Recent efforts on optical flow have primarily relied on deep neural networks.
FlowNet \cite{Dosovitskiy_2015_ICCV} first proposed the prototype of a CNN-based optical flow model. After that, a series of well-designed works were proposed, such as FlowNet2.0 \cite{Ilg_2017_CVPR}, SpyNet \cite{Ranjan_2017_CVPR}, and PWC-Net \cite{Sun_2018_CVPR,8621052}.
Then, the field made significant progress when RAFT \cite{10.1007/978-3-030-58536-5_24} proposed a new recurrent optical flow network to estimate optical flow.
Based on this breakthrough architecture, many recurrent networks \cite{Jiang_2021_CVPR,Luo_Yang_Luo_Li_Fan_Liu_2022,Sui_2022_CVPR,Xu_2021_ICCV,Zhang_2021_ICCV,Zheng_2022_CVPR,zhou2023samflow} have been proposed.
For example, GMA \cite{Jiang_2021_ICCV} suggested combining global motion to solve the problem of estimating occlusion, and KPA-Flow \cite{Luo_2022_CVPR}  designed kernel patch attention to deal with the local relationships of optical flow.
FlowFormer \cite{10.1007/978-3-031-19790-1_40} implemented a transformer structure in the optical flow network to capture long-range relations.
However, these methods do not focus on efficient inference and have heavy computational complexity due to their large number of iterations during inference. For example, the iteration number is 32 and 24 for Sintel \cite{10.1007/978-3-642-33783-3_44} and KITTI datasets \cite{Menze_2015_CVPR}, respectively.

\noindent
\textbf{Dynamic inference.}
As summarized by Han \emph{et al.} \cite{9560049}, dynamic inference networks have the advantages of efficiency, representation power, and interpretability since they can adapt the network structures during inference.
Inference has become more efficient in recent years with the help of sparse convolution \cite{Habibian_2021_CVPR,Parger_2022_CVPR,Wang_2021_CVPR,10.1007/978-3-030-58452-8_31,Yang_2022_CVPR}, early exiting strategies \cite{pmlr-v70-bolukbasi17a,huang2018multi,10.1007/978-3-031-19797-0_17,10.1007/978-3-030-58517-4_17}, and inference path selection \cite{Choi_2021_ICCV,Ding_2021_CVPR,Kong_2021_CVPR,Liu_2022_CVPR}.
For example, SMSR \cite{Wang_2021_CVPR} and QueryDet \cite{Yang_2022_CVPR} only use convolutions in important image areas and lower the computational cost of the unimportant region in super-resolution and object recognition tasks.
FrameExit \cite{Ghodrati_2021_CVPR} enables us to lower computational costs for video recognition by processing fewer frames for simpler videos. RBQE \cite{10.1007/978-3-030-58517-4_17} employs a faster process to remove minor artifacts for efficient compressed image enhancement.
ClassSR \cite{Kong_2021_CVPR} and MADA \cite{Choi_2021_ICCV} achieve efficient inference using a policy network to select different inference paths for each patch in super-resolution and video frame interpolation tasks.

\section{Methodology}

\subsection{Statistical Observation}
\label{sec:so}

\begin{figure}[t]
    \centering
    \includegraphics[width=\linewidth]{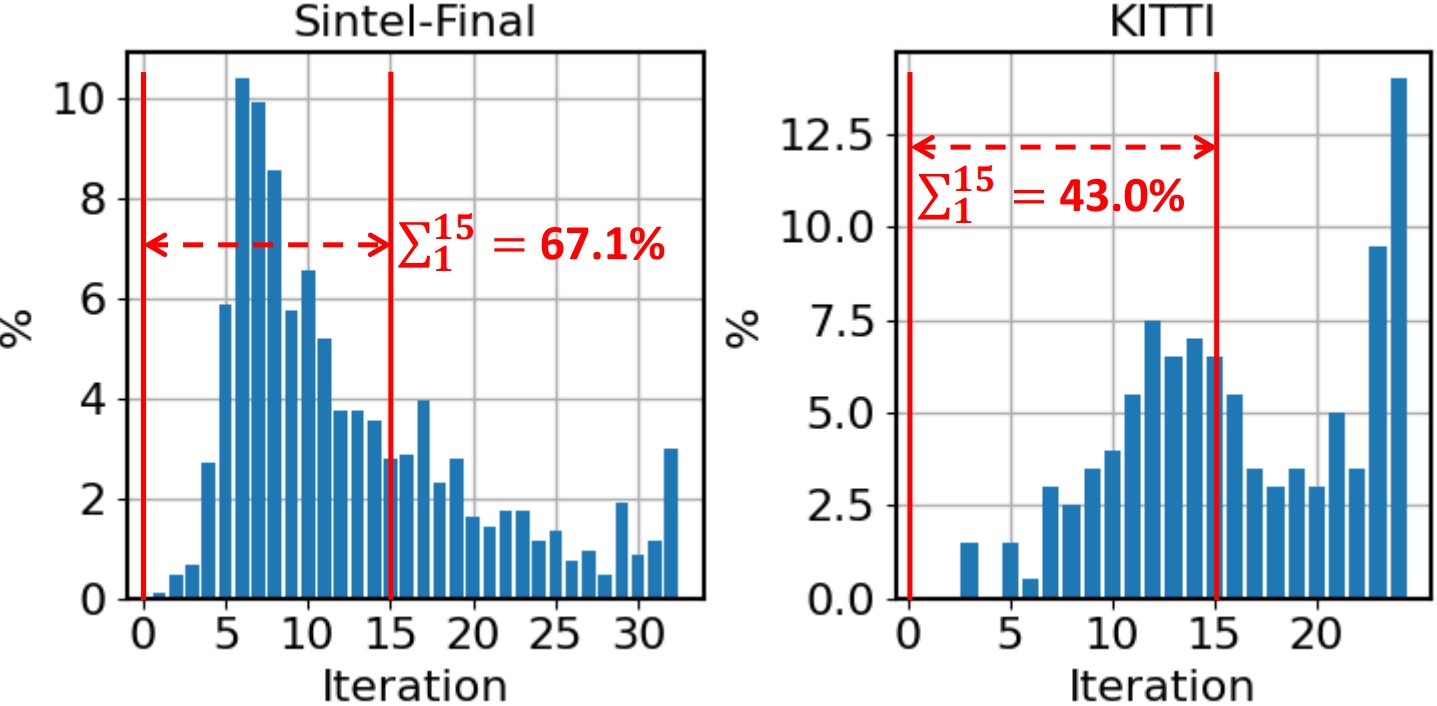}
    \caption{ \textbf{Statistical observation.} X-axis denotes the minimum number of iteration steps to achieve the near best EPE ($||EPE_x - EPE_{best}||<0.01$). The Y-axis denotes the percentage of samples that achieve the near EPE at the step.
    }
    \label{004_method}
    \end{figure}

In this subsection, we illustrate our statistic observation regarding the flow results of Sintel-Final and KITTI training datasets. These observations inspire us to design a more efficient dynamic optical flow network. 
We obtain the EPE value of each iteration step using RAFT \cite{10.1007/978-3-030-58536-5_24}.
Similar to recent works \cite{10.1007/978-3-030-58536-5_24,10.1007/978-3-031-19790-1_40,Luo_2022_CVPR}, the total iteration number of Sintel-Final and KITTI is 32 and 24, respectively.
Then we count the minimum number of iteration steps required to be within 0.01 of the best EPE for each sample and present the percentage of each iteration step in Figure~\ref{004_method}.

We can observe that approximately 67.1\% and 43.0\% of the samples within the Final and KITTI datasets, respectively, achieve almost the same results to the best EPE within the first 15 iterations.
This shows that the optical flow network hits a bottleneck, leading to $\Delta EPE$ close to zero. Therefore, we can skip the iterations to reduce computational complexity if the network encounters a bottleneck or the magnitude of EPE improvement is small. Since the information regarding the improvement of EPE is contained in previous iterations, we will incorporate the features of previous iterations into our policy network.
In addition, we can make the policy network perceive the magnitude of improvement in the future iteration to help it decide whether to enter the subsequent iteration.


    \begin{figure*}[t]
    \centering
    \includegraphics[width=\linewidth]{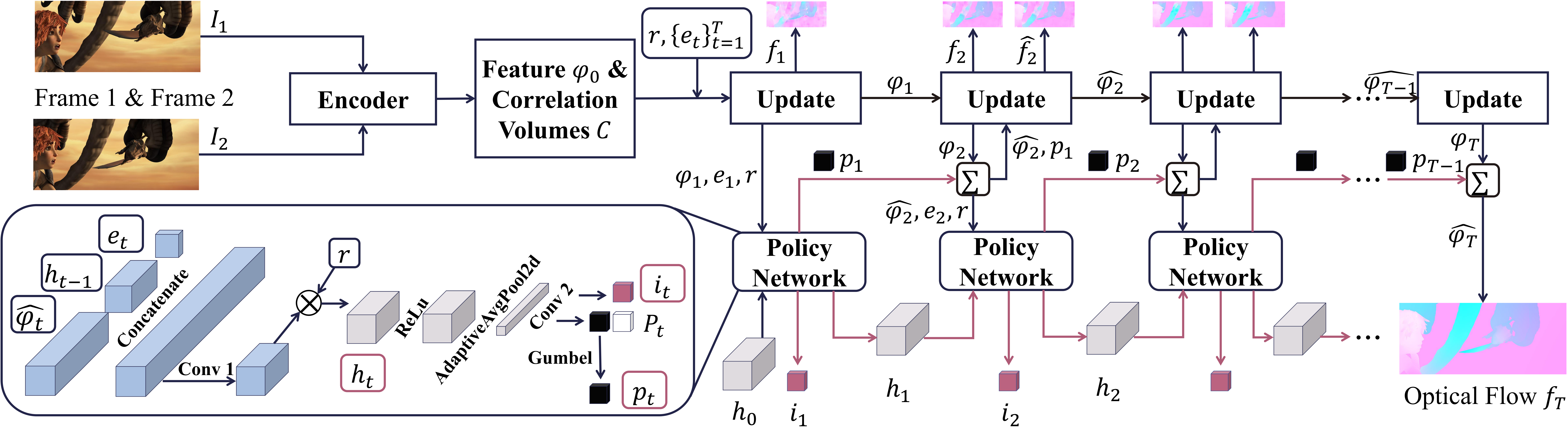}
    \caption{The architecture of the proposed dynamic optical flow network with the proposed context-aware iteration policy network. $\sum$ represents the aggregation described in Equation~\ref{equ:aggregate}, and we omit the aggregation for $\hat{f}_t$ in this figure. $\times$ denotes multiplication.
     }
    \label{005_method}
    \end{figure*}

\subsection{Overview of the Proposed Architecture}

Given the source image $I_1$ and the target image $I_2$, the task of optical flow estimation attempts to estimate a per-pixel displacement field between them.
Recurrent optical flow networks using deep learning have shown outstanding performance in recent years, and their structures typically include encoders for features and contexts, as well as an update operator. As shown in Figure~\ref{005_method}, the update operator takes the feature $\phi_{0}$ and a 4D cost volume $C$ as input and iteratively updates the flow $T$ times. The frame features $\phi_{0}$ come from the output of feature and context encoders, and the 4D cost volume is constructed from these frame features $\phi_{0}$.
In each iteration, the update operator generates a refined feature $\phi_{t}$ and optical flow $f_{t}$.
This recurrent update manner is similar to the steps of an optimization algorithm, and the optical flow estimation procedure is as follows:
\begin{equation}
\begin{aligned}
\phi_{0}, C &= \bm{Encoder} (I_1, I_2), \\
\phi_{t}, f_t &= \bm{Update} (\phi_{t-1}, f_{t-1}, C), t \in\{1,2,...,T \}.  \\
\end{aligned}
\end{equation}
The flow field $f_{0}$ is initialized to 0 everywhere.
Finally, the absolute distance between the ground truth flow $f_{gt}$ and the entire sequence of predictions ${f_1, ..., f_t}$ is used, which can be expressed as:
\begin{equation}
\mathcal{L}_{flow}(f_{gt}, \{{f}_t\}_1^T)=\sum_{t=1}^T \omega^{T-t}\left\|{f}_{g t}-{f}_t\right\|_1
\end{equation}
where $\omega$ is set to 0.8.

Our context-aware policy network can decide whether to skip the next iteration after each iteration performed by the update operator.
We use the iteration mask $p_t$ to indicate whether to skip or not , i.e., 0 for skip and 1 for enter.
To make the mask learnable, we use the Gumbel softmax trick \cite{jang2017categorical} to predict $p_t$ from the feature $P_t \in \mathbb{R}^{2 \times 1 \times 1}$ that is outputted by the last convolution in the policy network, which we express as follows:
\begin{equation}
p_t=\frac{\exp \left(\left({P}_{t}[0]+G_{t}[0]\right) / \tau\right)}{\sum_{i=0}^1 \exp \left(\left({P}_{t}[i]+G_{t}[i]\right) / \tau\right)},
\end{equation}
where $G_t$ is the Gumbel noise tensor which follows a Gumbel(0, 1) distribution, and $\tau$ is the temperature parameter. During inference, the network skips the update operator if $P_0 \textless P_1$ unless it enters the update operator.

Figure~\ref{003_intro} shows part of the inference procedure, where we see that the update operator does not bring any computational complexity if  we skip the update.
However, the update operator will execute all $T$ iterations during the training phase to make the network trainable, as shown in Figure~\ref{005_method}. 
We design the refined features and flow as an aggregation of previous and current outputs using the iteration mask. The aggregation is defined as follows:
\begin{equation}
\begin{aligned}
\hat{\phi}_{t} &= \phi_{t} \times p_{t-1} + \hat{\phi}_{t-1} \times (1 - p_{t-1}), \\ 
\hat{f}_{t} &= f_{t} \times p_{t-1} + \hat{f}_{t-1} \times (1 - p_{t-1}), \\
\end{aligned}
\label{equ:aggregate}
\end{equation}
where $t \in\{2,3,...,T \}$. If $p_{t-1}\rightarrow0$, the network skips the iteration, since $\hat{\phi}_{t}$ and $\hat{f}_{t}$ become $\hat{\phi}_{t-1}$ and $\hat{f}_{t-1}$. If $p_{t-1} \rightarrow 1$, the network enters the iteration, since $\hat{\phi}_{t}$ and $\hat{f}_{t}$ is $\phi_{t}$ and $f_{t}$.
Thus, the input of next update is $\hat{\phi}_{t},\hat{f}_{t}$ and $C$:
\begin{equation}
\begin{aligned}
\phi_{t+1}, f_{t+1} &= \bm{Update} (\hat{\phi}_{t},\hat{f}_{t}, C), t \in\{2,3,...,T-1 \}. \\
\end{aligned}
\label{equ:update}
\end{equation}

\subsection{Iteration Policy Network}
\noindent
\textbf{Controllable Network: }
 Our proposed dynamic optical flow network is user-controllable, allowing users to control its computational complexity based on their available computational resources.
We input the resource preference value $r$ into the policy network to achieve this. The policy network can condition on this resource preference value to determine whether to skip the subsequent iterations.
The smaller the resource preference value is, the greater the likelihood of skipping the next iteration, and vice versa. To introduce user resource preferences into the network, we multiply the output features of the first convolution in the policy network by the value $r$.
We then constraint the policy network output the iteration mask $p_t$ based on $r$ through the resource preference loss, which is expressed as follows:
\begin{equation}
\mathcal{L}_{res}= max(0,\frac{1}{T-1}\sum_{t=1}^{T-1} p_t - r).
\label{equ:res}
\end{equation}
$\mathcal{L}_{res}$ constraints the average $p_t$ should be smaller than resource preference $r$. As a result, we force the computational cost of the recurrent process to less than $r$ times the original.

\noindent
\textbf{Previous Iterations Information (`PI'): } Resource preference $r$ enables controllability of our dynamic network. However, the policy network is unable to accurately predict whether the flow improvement is limited using the features $\hat{\phi}_t$ in the iteration.
Based on the statistical observation discussed above, we input the previous iterations information from the history hidden cell $h_{t-1}$ and iteration embedding $e_t$ into our policy network, as shown in Figure~\ref{005_method}.
The encoding function for $e_t$ is $e_{t}=\{\sin \left(2^i \pi t\right), \cos \left(2^i \pi t\right)\}_{i=0}^2$.
$h_{t-1}$ and $e_t$ contains the information about flow change and progress information, assisting the policy network in determining whether the flow improvement has encountered a bottleneck.

 \begin{table*}[tb]
    \centering
    \resizebox{\textwidth}{!}
{
    \begin{tabular}{lclcclccclc}
    \toprule
     \multirow{2}{*}{Method} & \multicolumn{3}{c}{Sintel-Clean (train)} &  \multicolumn{3}{c}{Sintel-Final (train)} & \multicolumn{4}{c}{KITTI-15 (train)}  \\
    \cmidrule(r){2-4} \cmidrule(r){5-7} \cmidrule(r){8-11} & EPE & FLOPs(G) & Time(s) & EPE & FLOPs(G) & Time(s) & EPE & F1-all & FLOPs(G) & Time(s)  \\
    \hline
    \hline
    \multicolumn{11}{c}{C+T Training Data} \\
    \hline
    RAFT &  1.48 & 730 & 0.12 & 2.67 & 730  & 0.12 & 5.04 & 17.5 & 595 & 0.09 \\
    DRAFT & 1.48 &  \textcolor{red}{406(-44\%)} & \textcolor{red}{0.09} & 2.67 & \textcolor{red}{502(-31\%)} & \textcolor{red}{0.10} & 5.06 & 17.5 & \textcolor{red}{473(-21\%)} & \textcolor{red}{0.09} \\
    \hline
    GMA & 1.31 & 813 & 0.15 & 2.75 & 813 & 0.15 & 4.48 & 16.9 & 660 & 0.13 \\
    DGMA & 1.32 & \textcolor{red}{420(-48\%)} & \textcolor{red}{0.09} & 2.75 & \textcolor{red}{429(-47\%)} & \textcolor{red}{0.09} & 4.51 & 16.9 & \textcolor{red}{541(-18\%)} & \textcolor{red}{0.12}  \\
    \hline
    FlowFormer$^\ast$ & 0.94 & 974 & 0.31 & 2.33 & 974 & 0.31 & 4.10 & 14.5 & 496 & 0.19  \\
    DFlowFormer$^\ast$ & 0.94 & \textcolor{red}{572(-41\%)} & \textcolor{red}{0.21} & 2.33 & \textcolor{red}{496(-49\%)} & \textcolor{red}{0.20} & 4.11 & 14.5 & \textcolor{red}{403(-19\%)} & \textcolor{red}{0.17}  \\
    \hline
    KPA-Flow & 1.22 & 824 & 0.26 & 2.48 & 824 & 0.26 & 4.24 & 15.7 & 672 & 0.20 \\
    DKPA-Flow & 1.22 & \textcolor{red}{411(-50\%)} & \textcolor{red}{0.16} & 2.48 & \textcolor{red}{416(-49\%)} & \textcolor{red}{0.16} & 4.25 & 15.7 & \textcolor{red}{552(-18\%)} & \textcolor{red}{0.18} \\
    \hline
    \hline
    \multicolumn{11}{c}{C+T+S/K+(H) Training Data} \\
    \hline
    RAFT & (0.77) & 730 & 0.12 & (1.22) & 730 & 0.12 & (0.63) & (1.5) & 595 & 0.09 \\
    DRAFT & (0.77) & \textcolor{red}{283(-61\%)} & \textcolor{red}{0.06} & (1.22) & \textcolor{red}{301(-59\%)} & \textcolor{red}{0.07} & (0.63) & (1.5) & \textcolor{red}{248(-58\%)} & \textcolor{red}{0.05} \\
    \hline
    GMA &  (0.63) & 813 & 0.15 & (1.05) & 813 & 0.15 & (0.58) & (1.3) & 660 & 0.13 \\
    DGMA & (0.63) & \textcolor{red}{367(-55\%)} & \textcolor{red}{0.08} & (1.06) & \textcolor{red}{383(-53\%)} & \textcolor{red}{0.09} & (0.58) & (1.3) & \textcolor{red}{308(-53\%)} & \textcolor{red}{0.07} \\
    \hline
    
    FlowFormer$^\ast$ & (0.41) & 974 & 0.31 & (0.61) & 974 & 0.31 & (0.54) & (1.1) & 496 & 0.19 \\
    DFlowFormer$^\ast$ & (0.41) & \textcolor{red}{422(-57\%)} & \textcolor{red}{0.18} & (0.60) & \textcolor{red}{432(-56\%)} & \textcolor{red}{0.18} & (0.54) & (1.1) & \textcolor{red}{264(-47\%)} & \textcolor{red}{0.12} \\
    \hline
    KPA & (0.62) & 824 & 0.26 & (1.05) & 824 & 0.26 & (0.54) & (1.1) & 672 & 0.20  \\
    DKPA & (0.62) & \textcolor{red}{412(-50\%)} & \textcolor{red}{0.16}& (1.06) & \textcolor{red}{420(-49\%)}  & \textcolor{red}{0.16} & (0.54) & (1.1) & \textcolor{red}{286(-57\%)} & \textcolor{red}{0.11} \\
\toprule
    \end{tabular}
}
    \caption{Quantitative comparison on Sintel and KITTI 2015 training datasets. $\downarrow$EPE/$\downarrow$F1-all/$\downarrow$FLOPs(G)/$\downarrow$Time(s) are used for evaluation.  `C+T' refers to results that are trained on Chairs \cite{Dosovitskiy_2015_ICCV} and Things \cite{Mayer_2016_CVPR} datasets.
“S/K(+H)” refers to methods fine-tuned on Sintel \cite{10.1007/978-3-642-33783-3_44}, KITTI \cite{Menze_2015_CVPR}, and some on HD1K \cite{Kondermann_2016_CVPR_Workshops} datasets. The red text denotes the best result, and parentheses indicate the training results. $\ast$ denotes that FlowFormer should forward four times to obtain the optical flow, and here we show the FLOPs and time for one forward. As described in their paper, transformers are sensitive to image size, so they crop the image into four pieces and feed it into the model four times in their code.
}
    \label{table:compare-train}
\end{table*}

Specifically, iteration embedding indicates the iteration progress information, which informs the policy network the position and order of the current iteration.
In addition, the historical hidden cell is the condensed features of flow, iteration embedding and iteration mask in the previous iteration, so it introduces the previous policy decision and flow change information to policy network. 
By convolving the previous decision information in $h_{t-1}$ with the current feature $\hat{\phi}_t$ and $e_t$, the policy network can determine whether the optical flow has been improved in the current iteration.
The entire process of obtaining $h_t$ and $p_t$ is defined as follows:
\begin{equation}
\begin{aligned}
h_t &=Conv_1 (Concatenate\{\hat{\phi}_{t}, h_{t-1}, e_{t}\}) \times r, \\
P_t, i_t &=Conv_2(AdaptiveAvgPool(ReLu(h_t))), \\
p_t &=Gumbel(P_t), \\
\end{aligned}
\end{equation}
where $t \in\{1,2,...,T-1 \}$, and $h_0$ is initialized to 0 everywhere. We obtain the history hidden cell $h_{t}$ by multiplying the resource preference value $r$ with the output of first convolution.  Then, we input the $h_{t}$ through a ReLu activation to enhance the ability of network representation and an AdaptiveAvgPool operator for the fusion of global feature. Finally, we output the $P_t$ and $i_t$ by the second convolution, and we obtain $p_t$ by inputting $P_t$ into the Gumbel softmax.
$i_t$ is the predicted magnitude of flow improvement, which we will describe how it works later.
The computational cost of the policy network is negligible compared with the update operator since the FLOPs of the policy network is less than 1\% of the update operator in RAFT.
The procedure of policy network can be summarized as follows:
\begin{equation}
\begin{aligned}
h_t, p_t, i_t &=\bm{Policy} (\hat{\phi}_{t}, h_{t-1}, e_{t}, r), t \in\{1,2,...,T-1 \}, \\
\end{aligned}
\end{equation}
where $\hat{\phi}_{1}$ is set to ${\phi}_{0}$.

\noindent
\textbf{Future Iterations Information (`FI'): }
The magnitude of flow improvement in subsequent iterations is important for the policy network to make decisions based on the statistical observation discussed above.
The efficient networks should dedicate computational resources to iterations where samples can have a significant optical flow improvement.
Therefore, we use an incremental loss to make our policy network predict the flow improvement $i_t$ in subsequent iterations. The incremental loss $\mathcal{L}_{incre}$ is expressed as follows:
\begin{equation}
\mathcal{L}_{incre}=\sum_{t=1}^{T-1} \left\| \|{f}_{g t}-\hat{f}_t\|_1 -\|{f}_{g t}-{f}_{t+1}\|_1 -i_t \right\|_1,
\end{equation}
where $\hat{f}_t$ is the output flow of the current iteration after the aggregate operation, as described in Equation~\ref{equ:aggregate}, and ${f}_{t+1}$ is the output flow of the next iteration before the aggregate operation.
Predicting future improvements is difficult, but our network acquires the information about the approximate magnitude improvements in optical flow.
As shown in Figure~\ref{005_method}, since the predictions of iteration mask $p_t$ and incremental improvement $i_t$ are based on the same features, the estimation of $p_t$ can implicitly refer to information about the magnitude of optical flow improvement.

\subsection{Overall Loss}
\label{sec:olf}
We summarize the training loss described above as follows:
\begin{equation}
        \mathcal{L}_{{overall }}=\mathcal{L}_{flow}(f_{gt}, \{\hat{f}_t\}_1^T) +\lambda_{res}\mathcal{L}_{res} +\lambda_{incre} \mathcal{L}_{incre},
\label{equ:overall}
\end{equation}
where $\lambda_{res}$ and $\lambda_{incre}$ are the weight for $\mathcal{L}_{res}$ and $\mathcal{L}_{incre}$.

\section{Experiment}

Our iteration policy network and training strategy can seamlessly integrate into contemporary optical flow architecture, and the policy network is FLOPs-efficient.
In this work, we select four state-of-the-art methods, including RAFT \cite{10.1007/978-3-030-58536-5_24}, GMA \cite{Jiang_2021_ICCV}, FlowFormer \cite{10.1007/978-3-031-19790-1_40}, and KPA \cite{Luo_2022_CVPR}, as our backbone.
They are recurrent-based deep learning optical flow networks, which iteratively update the flow field through an update operator.

\subsection{Evaluation Metrics}




We measure average Endpoint Error (EPE) and the percentage of optical flow outliers over all pixels (F1-all) for Sintel \cite{10.1007/978-3-642-33783-3_44} and KITTI \cite{Menze_2015_CVPR} datasets.
$EPE = \left\|f_{g t}-f_t\right\|_2$, and 
\textit{F1-all}$=\frac{EPE>3 \text { and } EPE/{\|f_t\|}>0.05}{\# \text { valid pixels }}$.
Outlier is the pixel satisfying ${EPE>3 \text { and } EPE/{\|f_t\|}>0.05}$.
Each method was evaluated on an NVIDIA GeForce RTX 3090 GPU to measure the inference speed per sample.
Additionally, we count floating point operations (FLOPs) to determine the computational complexity by running RAFT, GMA, and KPA-Flow at a resolution of 1248$\times$376 for Sintel and 1024$\times$440 for KITTI, respectively.
For FlowFormer, the resolution for KITTI training, KITTI test, and Sintel datasets is 720$\times$376, 1242$\times$432, and 960$\times$432, as they provide in the code.

\subsection{Implementation Details}

The iteration policy network is implemented as shown in Figure~\ref{005_method}, which FLOPs is less than 1\% of the update operator in RAFT.
We used the pre-trained network to initialize the network and fixed the network except for our policy network during training.

Specifically, we initialized the network parameter for RAFT and KPA-Flow with the pre-trained network trained on the FlyingThings \cite{Mayer_2016_CVPR} dataset and then trained them for 40k iterations. 
This model was used to evaluate the Sintel and  KITTI training datasets.
To obtain models for the Sintel and  KITTI test datasets, we initialized network parameters with the pre-trained network trained on Sintel, KITTI, and HD1K \cite{Kondermann_2016_CVPR_Workshops}.
Then, we trained the policy network parameters using 40k and 20k iterations for Sintel and KITTI, respectively.
For GMA and FlowFormer, we trained models for Sintel and KITTI training datasets and Sintel test datasets using 50k iterations.
In addition, like recent works, the total testing iterations for Sintel and KITTI are 32 and 24, while the training iteration is 12.
The weights $\lambda_{res}$ and $\lambda_{incre}$ in the overall loss (Equation~\ref{equ:overall}) are set to 50 and 1. $r$ is randomly sampled from 0.2 $\sim$ 1.0. The learning rate is the same with their codes.

\subsection{Policy Network with Existing Flow Networks}
Using our proposed iteration policy network, RAFT, GMA, FlowFormer, and KPA-Flow are denoted as DRAFT, DGMA, DFlowFormer, and DKPA-Flow, respectively. Table~\ref{table:compare-train} and Table~\ref{table:compare-test} show that our dynamic network can maintain performance while achieving lower computational cost, with approximately 40\%/20\% reduction in FLOPs for Sintel/KITTI datasets.
For example, Table~\ref{table:compare-train} displays that the F1-all metric of all our dynamic networks is consistent with the origin backbone for the KITTI datasets. However, the FLOPs is reduced by around 20\% and 50\% for our dynamic networks trained on C+T and C+T+S/K+(H), respectively.
In addition,  our dynamic networks maintain the performance on Sintel-Final, but our models (DRAFT, DGMA, DFlowFormer, DKPA-Flow) reduce the FLOPs by 43\%/43\%/23\%/35\%, as shown in the Table~\ref{table:compare-test}.

    \begin{figure}[t]
    \centering
    \includegraphics[width=\linewidth]{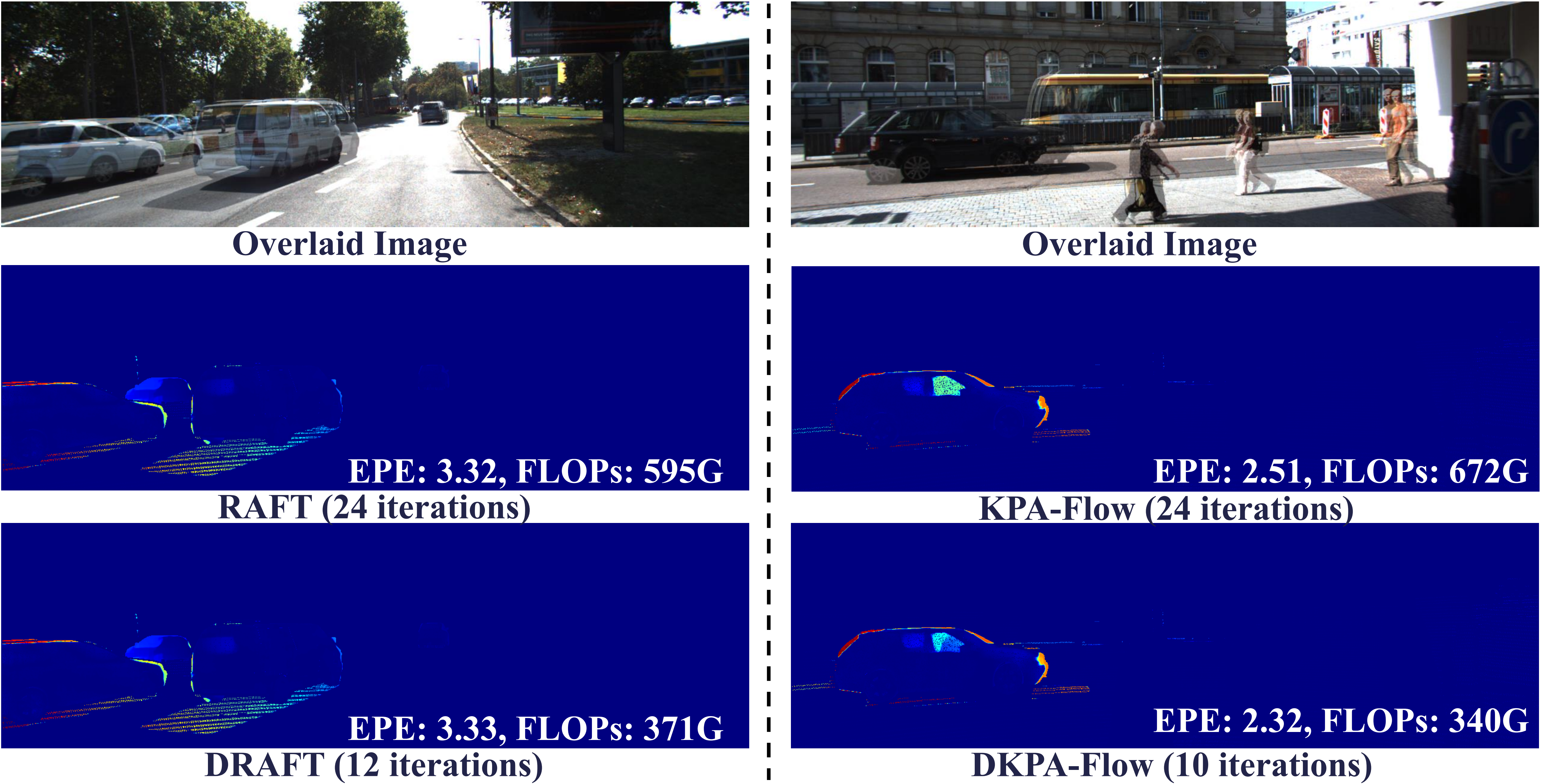}
    \caption{Qualitative comparison on the KITTI-train dataset. In the EPE map, the blue color is better, and the red is worse.
    }
    \label{009_compare}
    \end{figure}


In addition, we display the visualization results of EPE map in Figure~\ref{009_compare}, illustrating that our dynamic models require fewer FLOPs to estimate the optical flow, which is comparable or better to utilizing the original backbone.



\subsection{Ablation Study}
For each ablation model, we increase its FLOPs to exceed our DRAFT or DFlowFormer to verify the effectiveness of our proposed approaches. 

\begin{figure}[t]
    \centering
    \includegraphics[width=\linewidth]{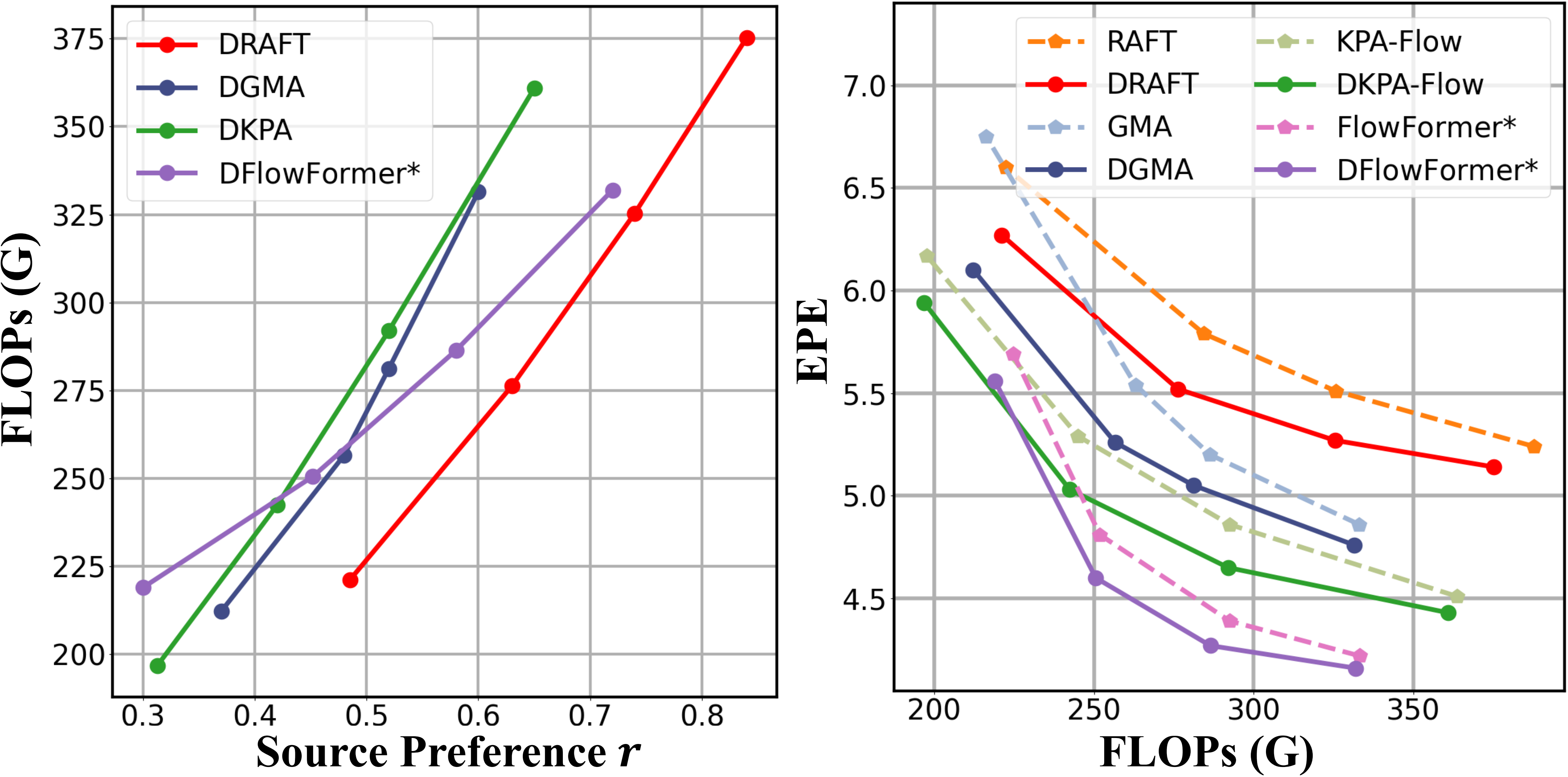}
    \caption{Ablation study of the controllable computational complexity on KITTI. We plot the graph of FLOPs changing with source preference $r$ and the graph of EPE changing with FLOPs.
    }
    \label{006_ablation}
    \end{figure}

\noindent
\textbf{Controllable Computational Complexity.}
Users can change their resource preference $r$ to control the number of FLOPs, which is an excellent property of our dynamic network.
For example, Figure~\ref{006_ablation} illustrates that when we increase the $r$ from 0.5 to 0.8, the FLOPs(G) of DRAFT also increases from 225 to 375. In addition, FLOPs and $r$ are close to a proportional relationship, which is convenient for users to control FLOPs.

 \begin{table*}[tb]
    \centering
    \resizebox{\textwidth}{!}
{
    \begin{tabular}{lclcclcclc}
    \toprule
    \multirow{2}{*}{Method} & \multicolumn{3}{c}{Sintel-Clean (test)} &  \multicolumn{3}{c}{Sintel-Final (test)} & \multicolumn{3}{c}{KITTI-15 (test)}  \\
    \cmidrule(r){2-4} \cmidrule(r){5-7} \cmidrule(r){8-10} & EPE & FLOPs(G) & Time(s) & EPE & FLOPs(G) & Time(s) & F1-all & FLOPs(G) & Time(s)  \\
    \hline
    \hline
    RAFT & 1.94 & 730 & 0.12 & 3.18 & 730 & 0.12 & 5.10  & 595 & 0.09 \\
    DRAFT & 1.93 & \textcolor{red}{411(-44\%)} & \textcolor{red}{0.09}  & 3.20 & \textcolor{red}{412(-43\%)} & \textcolor{red}{0.09}  & 5.16 &  \textcolor{red}{465(-22\%)} & \textcolor{red}{0.09} \\
    \hline
    GMA & 1.39 & 813 & 0.15 & 2.47 & 813 & 0.15 & 5.15  & 660 & 0.13 \\
    DGMA & 1.43 & \textcolor{red}{469(-42\%)} & \textcolor{red}{0.10} & 2.53 & \textcolor{red}{465(-43\%)} & \textcolor{red}{0.10} & 5.18  & \textcolor{red}{507(-23\%)} & \textcolor{red}{0.10}  \\
\hline
    FlowFormer$^\ast$ & 1.15 & 974 & 0.31 & 2.18 & 974 & 0.31 & 4.70 & 808 & 0.29 \\
    DFlowFormer$^\ast$ & 1.17 & \textcolor{red}{627(-36\%)} & \textcolor{red}{0.24} & 2.21 & \textcolor{red}{752(-23\%)} & \textcolor{red}{0.26} & 4.74 & \textcolor{red}{689(-15\%)} & \textcolor{red}{0.26} \\

\hline
    KPA-Flow & 1.35 & 824 & 0.26 & 2.27 & 824 & 0.26 & 4.67  & 672 & 0.20 \\
    DKPA-Flow & 1.35 & \textcolor{red}{527(-36\%)} & \textcolor{red}{0.21} & 2.29 & \textcolor{red}{533(-35\%)} & \textcolor{red}{0.21} &  4.69 & \textcolor{red}{519(-23\%)} & \textcolor{red}{0.18} \\
    \toprule
    \end{tabular}
}
\caption{Quantitative comparison on Sintel and KITTI test datasets. All models are trained on C+T+S/K+(H) training data.}
    \label{table:compare-test}
\end{table*} 


Figure~\ref{006_ablation} also shows that EPE improves with FLOPs, and our dynamic networks outperform the original backbone when their FLOPs is close. For example, when the FLOPs is around 325G, the EPE of our DRAFT is 5.27, which is better than 5.51 EPE of RAFT.




        \begin{table}[tb]
        \centering
        \small
        \resizebox{\linewidth}{!}
    {   
        \begin{tabular}{lcccccccccc}
        \toprule
        \multirow{2}{*}{Method} & \multirow{2}{*}{$\mathcal{L}_{res}$} & \multicolumn{2}{c}{Information} & \multicolumn{1}{c}{Clean} &  \multicolumn{1}{c}{Final} & \multicolumn{2}{c}{KITTI}  \\
        \cmidrule(r){3-4} \cmidrule(r){5-5} \cmidrule(r){6-6} \cmidrule(r){7-8} & & PI & FI & EPE & EPE & EPE & F1-all \\
        \hline
        DRAFT-$\mathcal{L}_{1}$ & - & \Checkmark & \Checkmark & 1.73 & 2.97 & 5.84 & 19.2 \\
        DRAFT-B & \Checkmark & - & - & 1.81 & 2.99 & 6.97 & 25.7 \\
        DRAFT-P & \Checkmark & \Checkmark & - & 1.58 & 2.77 & 5.66 & 19.0 \\
        DRAFT & \Checkmark & \Checkmark & \Checkmark & \textcolor{red}{1.56} & \textcolor{red}{2.74} & \textcolor{red}{5.52} & \textcolor{red}{19.0}  \\
        \hline
        DFlowFormer-$\mathcal{L}_{1}$ & - & \Checkmark & \Checkmark & 1.15 & 2.60 & 4.30 & 15.1  \\
        DFlowFormer-B & \Checkmark & - & - & 1.20 & 2.65 & 4.84 & 19.4  \\
        DFlowFormer-P & \Checkmark & \Checkmark & - & 1.07 & 2.51 & 4.23 & 15.2  \\
        DFlowFormer & \Checkmark & \Checkmark & \Checkmark & \textcolor{red}{1.03} & \textcolor{red}{2.43} & \textcolor{red}{4.21} & \textcolor{red}{15.1}  \\
        \toprule
        \end{tabular}
    }
        \caption{Quantitative comparison of the $\mathcal{L}_{res}$ and contextual information ablation study. `-$\mathcal{L}_{1}$' denotes the model uses absolute distance to control the computational cost.
         Previous iterations information (`PI') includes historical hidden cell $h_{t-1}$ and iteration embedding $e_t$ in the policy network as described in Eqa.(7). 
        Future iterations information (`FI') is the flow improvement $i_t$ in subsequent iterations estimated by the network and constrained by an incremental loss $\mathcal{L}_{incre}$.
        }
        \label{table:ablation_info}
    \end{table}

    \begin{figure}[t]
    \centering
    \includegraphics[width=\linewidth]{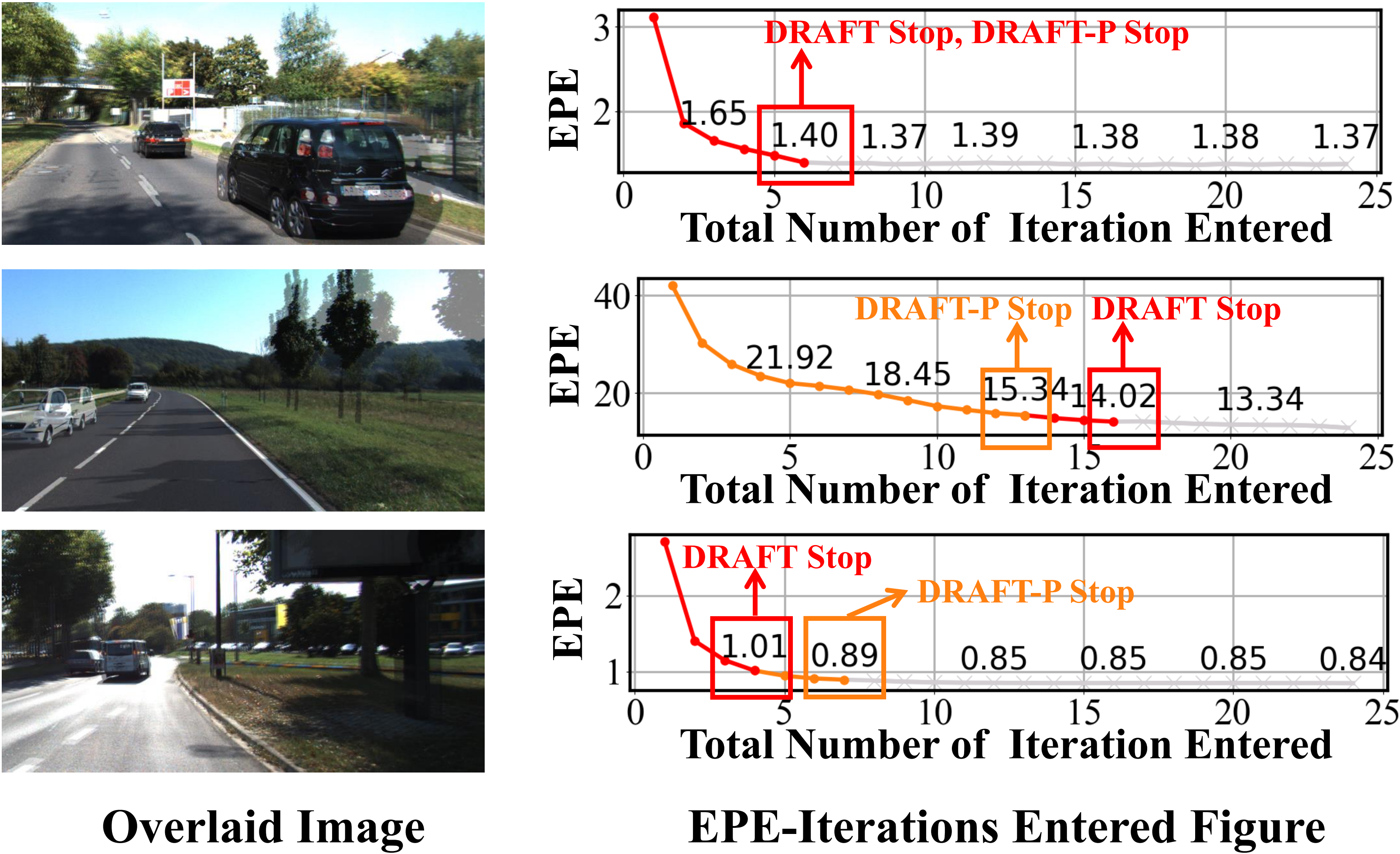}
    \caption{
    Three visualization examples of the contextual information ablation study on KITTI.
    }
    \label{007_ablation}
    \end{figure}

\noindent
\textbf{Ablation Study of $\mathcal{L}_{res}$.}
$\mathcal{L}_{res}$ in Equation~\ref{equ:res} uses the $max(0,x)$, forcing the computational cost of the recurrent process to be less than $r$ times the original. 
We implement an ablation model denoted as `-$\mathcal{L}_{1}$,' which uses absolute distance $\mathcal{L}_{1}$ instead of $max(0,x)$.
Specifically, $\mathcal{L}_{1}$ we used is: $\mathcal{L}_{1}=\|\frac{1}{T-1}\sum_{t=1}^{T-1} p_t - r\|_1$.
Table~\ref{table:ablation_info} shows that `-$\mathcal{L}_{1}$' ablation models perform poorly. For example, DRAFT reduces the EPE of Clean/Final/KITTI by 0.17/0.23/0.32 compared to DRAFT-$\mathcal{L}_{1}$. 


\noindent
\textbf{Ablation Study of Contextual Information.}
We implement two ablation models, `-B' denotes the policy network does not rely on contextual information, and `-P' denotes the policy network only utilizes previous iterations information.
Table~\ref{table:ablation_info} shows the `-B' ablation model performs worst among all models. For example, DRAFT-B has 0.25/0.22/1.45 more EPE than DRAFT for Clean/Final/KITTI. 
We find that the optimal number of iterations predicted by the `-B' model is either extremely small or extremely large, indicating that the network is incapable of making accurate predictions in the absence of contextual information.
The '-P' ablation model helps decide by feeding historical hidden cell and iteration embedding into the policy network, but its performance is still inferior to DRAFT or DFlowFormer, which exploits contextual information. For example, Clean/Final/KITTI EPE is reduced by 0.02/0.03/0.14 when comparing DRAFT to DRAFT-P.

To further understand how contextual information works, we provide three visualization examples in Figure~\ref{007_ablation}.
By incorporating previous information into our policy network, the policy network knows whether the flow improvement has hit a bottleneck. As shown in the first example of Figure~\ref{007_ablation}, DRAFT and DRAFT-P only enter iteration 6 times since the improvement of optical flow after that point is negligible.


From the second and third examples in Figure~\ref{007_ablation}(b), we can observe that DRAFT skips 3 more iterations for the third example, so EPE increases the EPE by 0.12 compared to DRAFT-P.
However, DRAFT reduces 1.32 EPE for the second example by entering 3 more iterations. These observations indicate that by using incremental loss to help the policy network perceive future information, the policy network can assign more iterations to samples with a large improvement in subsequent iterations.

\section{Conclusion}
This paper proposes a novel context-aware iteration policy network for efficient optical flow estimation.
We find that determining whether flow improvement is bottlenecked or minimal is crucial for reducing the computational cost.
Therefore, historical hidden cell and iteration embedding are introduced to provide information about previous iterations, allowing the policy network to determine whether the flow improvement hit a bottleneck. In addition, the policy network assigns more iterations to the sample with large improvement in the future iteration by using the incremental loss. 
The ablation study shows the controllability and effectiveness of our FLOPs control strategy and the usefulness of contextual information.
Extensive experiments demonstrate that our network maintains the performance with a 40\%/20\% reduction in FLOPs for the Sintel/KITTI datasets.

\bibliography{aaai24}

\newpage

\appendix

\begin{algorithm}[t]
\small
	\caption{Training phase of our dynamic optical flow network.}
    
	\KwIn{Initialize network parameters $\theta$  of dynamic optical flow network. Initialize iteration embedding $\{e_t\}_{t=1}^{T-1}$. Set $\lambda_{res}$ to 50 and $\lambda_{incre}$ to 1.
 }
	\BlankLine

        
	\While{\textnormal{$\theta$ has not converged}}{
    	 Sample $\{I_1, I_2\}$ a batch from the dataset; \\
        $\{\phi_{0}, C\} \leftarrow \bm{Encoder} (I_1, I_2)$; \\
        $\{f_0,h_0\} \leftarrow \{ \bm{0}, \bm{0} \}$; \\
        $\{r\} \leftarrow Random(0.2, 1.0)$; \\
        $\{\phi_{1}, f_1\} \leftarrow \bm{Update} ({\phi}_{0},{f}_{0}, C)$; \\
        \{$h_1, p_1, i_1\} \leftarrow \bm{Policy} ({\phi}_{1}, h_{0}, e_{1}, r)$; \\
        $\{\hat{\phi}_1,\hat{f}_1\} \leftarrow \{ \phi_{1}, f_1 \}$; \\
        \ForEach{ $t$ in \{2, 3, ..., T-1 \} }{
    	$\{\phi_{t}, f_t\} \leftarrow \bm{Update} (\hat{\phi}_{t-1},\hat{f}_{t-1}, C)$; \\
            $\hat{\phi}_{t} \leftarrow \phi_{t} \times p_{t-1} + \hat{\phi}_{t-1} \times (1 - p_{t-1})$; \\ 
            $\hat{f}_{t} \leftarrow f_{t} \times p_{t-1} + \hat{f}_{t-1} \times (1 - p_{t-1})$; \\ 
            \{$h_t, p_t, i_t\} \leftarrow \bm{Policy} (\hat{\phi}_{t}, h_{t-1}, e_{t}, r)$; \\
		}
      $\{\phi_{T}, f_T\} \leftarrow \bm{Update} (\hat{\phi}_{T-1},\hat{f}_{T-1}, C)$; \\
    $\hat{f}_{T} \leftarrow f_{T} \times p_{T-1} + \hat{f}_{T-1} \times (1 - p_{T-1})$; \\ 
          $\mathcal{L}_{{overall }} \leftarrow  \mathcal{L}_{flow}( f_{gt}, \{\hat{f}_t\}_1^T) +\lambda_{res}\mathcal{L}_{res} +\lambda_{incre} \mathcal{L}_{incre}$; \\
        $\theta \leftarrow \theta- \nabla_{\theta} \mathcal{L}_{overall}$;\\
	}
    \BlankLine
    \KwOut{Dynamic optical flow network.}  
	\label{alg:dofn_train}
\end{algorithm}

\begin{algorithm}[t]
\small
	\caption{Inference phase of our dynamic optical flow network.}
    
	\KwIn{Dynamic optical flow network trained using the Algorithm~\ref{alg:dofn_train}. Initialize iteration embedding $\{e_t\}_{t=1}^{T-1}$. Set resource preference value $r$ based on the computational resource.
 }
	\BlankLine

        
	\While{\textnormal{$\theta$ has not converged}}{
    	 Sample $\{I_1, I_2\}$ a batch from the dataset; \\
        $\{\phi_{0}, C\} \leftarrow \bm{Encoder} (I_1, I_2)$; \\
        $\{f_0,h_0\} \leftarrow \{ \bm{0}, \bm{0} \}$; \\
        $\{\phi_{1}, f_1\} \leftarrow \bm{Update} ({\phi}_{0},{f}_{0}, C)$; \\
        \{$h_1, p_1, i_1\} \leftarrow \bm{Policy} ({\phi}_{1}, h_{0}, e_{1}, r)$; \\
        $\{\hat{\phi}_1,\hat{f}_1\} \leftarrow \{ \phi_{1}, f_1 \}$; \\
        \ForEach{ $t$ in \{2, 3, ..., T-1 \} }{
        $ \{ \phi_{t}, f_t \} \leftarrow \{\hat{\phi}_{t-1},\hat{f}_{t-1}\} $; \\
            \If{$P_0 \textgreater P_1 $}{
                $\{\phi_{t}, f_t\} \leftarrow \bm{Update} (\hat{\phi}_{t-1},\hat{f}_{t-1}, C)$; \\
            }
        $\{\hat{\phi}_t,\hat{f}_t\} \leftarrow \{ \phi_{t}, f_t \}$; \\
            \{$h_t, p_t, i_t\} \leftarrow \bm{Policy} (\hat{\phi}_{t}, h_{t-1}, e_{t}, r)$; \\
		}
          $ \{ f_{T} \} \leftarrow \{\hat{f}_{T-1}\} $; \\
  \If{$P_0 \textgreater P_1 $}{
      $\{\phi_{T}, f_T\} \leftarrow \bm{Update} (\hat{\phi}_{T-1},\hat{f}_{T-1}, C)$; \\
            }
	}
    \BlankLine
    \KwOut{$f_T$}  
	\label{alg:dofn_test}
\end{algorithm}

    \begin{table*}[t]
        \centering
        \small
        \begin{tabular}{lccccccc}
        \hline
        \multirow{2}{*}{Method} & \multicolumn{2}{c}{Sintel-Clean (train)} &  \multicolumn{2}{c}{Sintel-Final (train)} & \multicolumn{3}{c}{KITTI-15 (train)}  \\
        \cmidrule(r){2-3} \cmidrule(r){4-5} \cmidrule(r){6-8}  & EPE & FLOPs(G) & EPE & FLOPs(G) & EPE & F1-all & FLOPs(G)  \\
        \hline
        DRAFT-Exit & 1.57 & 198.41 & 2.76 & 223.79  & 5.55 & 19.4 & 276.35 \\
        DRAFT-$\mathcal{L}_{1}$ & 1.73 & 207.61 & 2.97 & 227.49 & 5.84 & 19.2 & 283.60 \\
        DRAFT-B & 1.81 &  199.34 & 2.99 & 223.54 & 6.97 & 25.7 & 277.28 \\
        DRAFT-P  & 1.58 & 197.92  & 2.77 & 223.96 & 5.66 & 19.0 & 276.70 \\
        DRAFT & \textcolor{red}{1.56} & \textcolor{red}{197.27}  & \textcolor{red}{2.74} & \textcolor{red}{223.45}  & \textcolor{red}{5.52} & \textcolor{red}{19.0} & \textcolor{red}{276.26} \\
        \hline
        \hline
        DFlowformer-Exit & 1.04 & 381.39 & 2.45 & 390.83 & 4.35 & 16.2 & 306.34  \\
        DFlowformer-$\mathcal{L}_{1}$ & 1.15 & 385.42 & 2.60 & 392.03 & 4.30 & 15.1 & 313.86  \\
        DFlowformer-B & 1.20 & 399.41 & 2.65 & 393.22 & 4.84 & 19.4 & 331.23  \\
        DFlowformer-P & 1.07 & 380.88 & 2.51 & 391.31 & 4.23 & 15.2 & 304.48  \\
        DFlowformer & \textcolor{red}{1.03} & \textcolor{red}{380.30} & \textcolor{red}{2.43} & \textcolor{red}{390.51} & \textcolor{red}{4.21} & \textcolor{red}{15.1} & \textcolor{red}{304.48}  \\
        \toprule
        \end{tabular}
        \caption{FLOPs of ablation models. For each ablation model, we increase its FLOPs to slightly exceed our dynamic network DRAFT or DFlowformer to verify the effectiveness of our proposed approaches.}
        \label{table:ablation_s}
    \end{table*}

    \begin{figure}[t]
    \centering
    \includegraphics[width=0.9\linewidth]{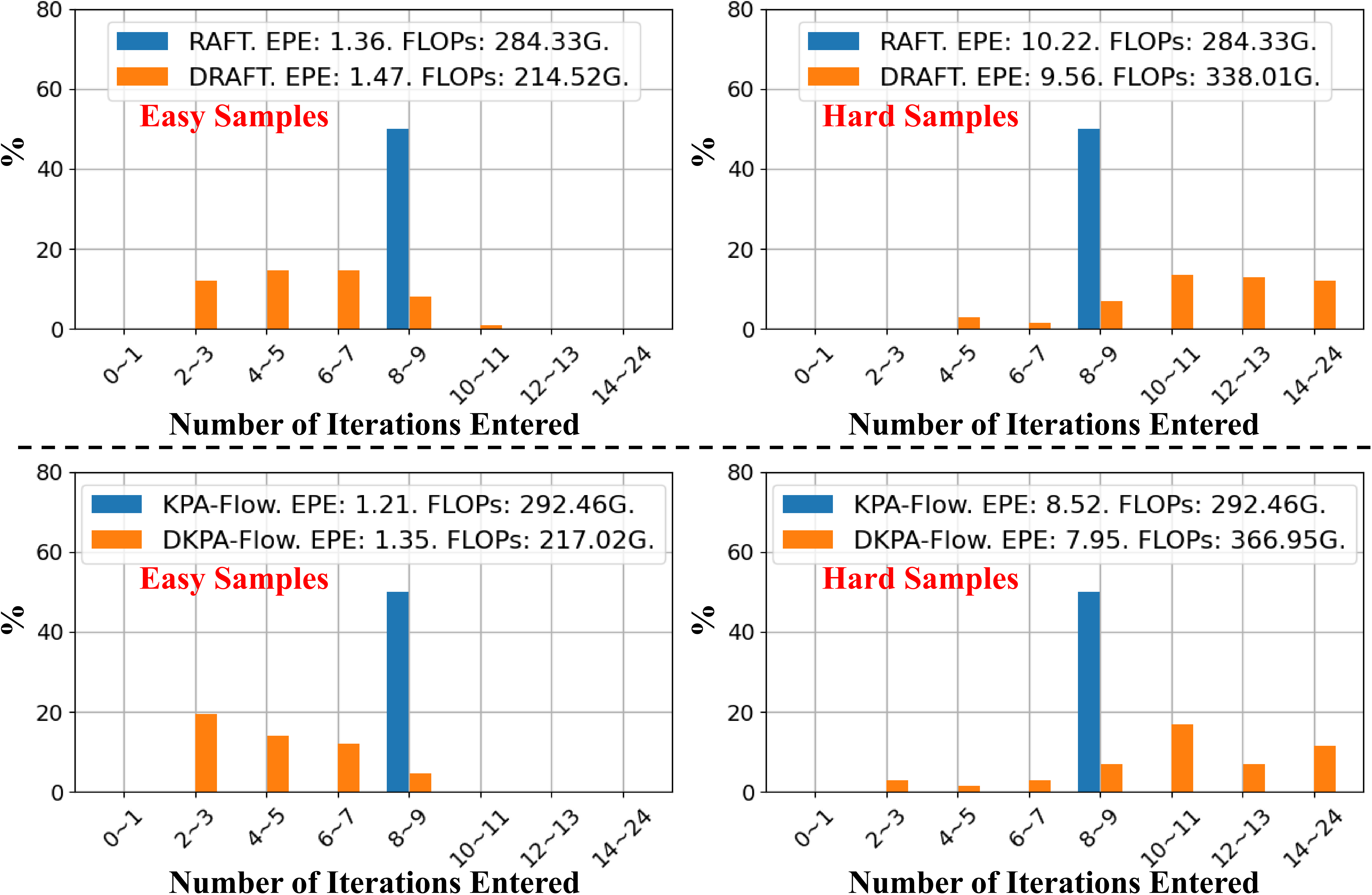}
    \caption{The comparison of the entering iteration number with RAFT and KPA-Flow on KITTI. We reduce the number of iterations of RAFT and KPA-Flow to compare the performance at almost the same FLOPs fairly. The FLOPs of all samples for DRAFT and DKPA-Flow are 276.26G and 291.99G, which are lower than RAFT (284.33G) and KPA-Flow (292.46G).
    }
    \label{011_compare_s}
    \end{figure}

    \begin{figure}[t]
    \centering
    \includegraphics[width=0.9\linewidth]{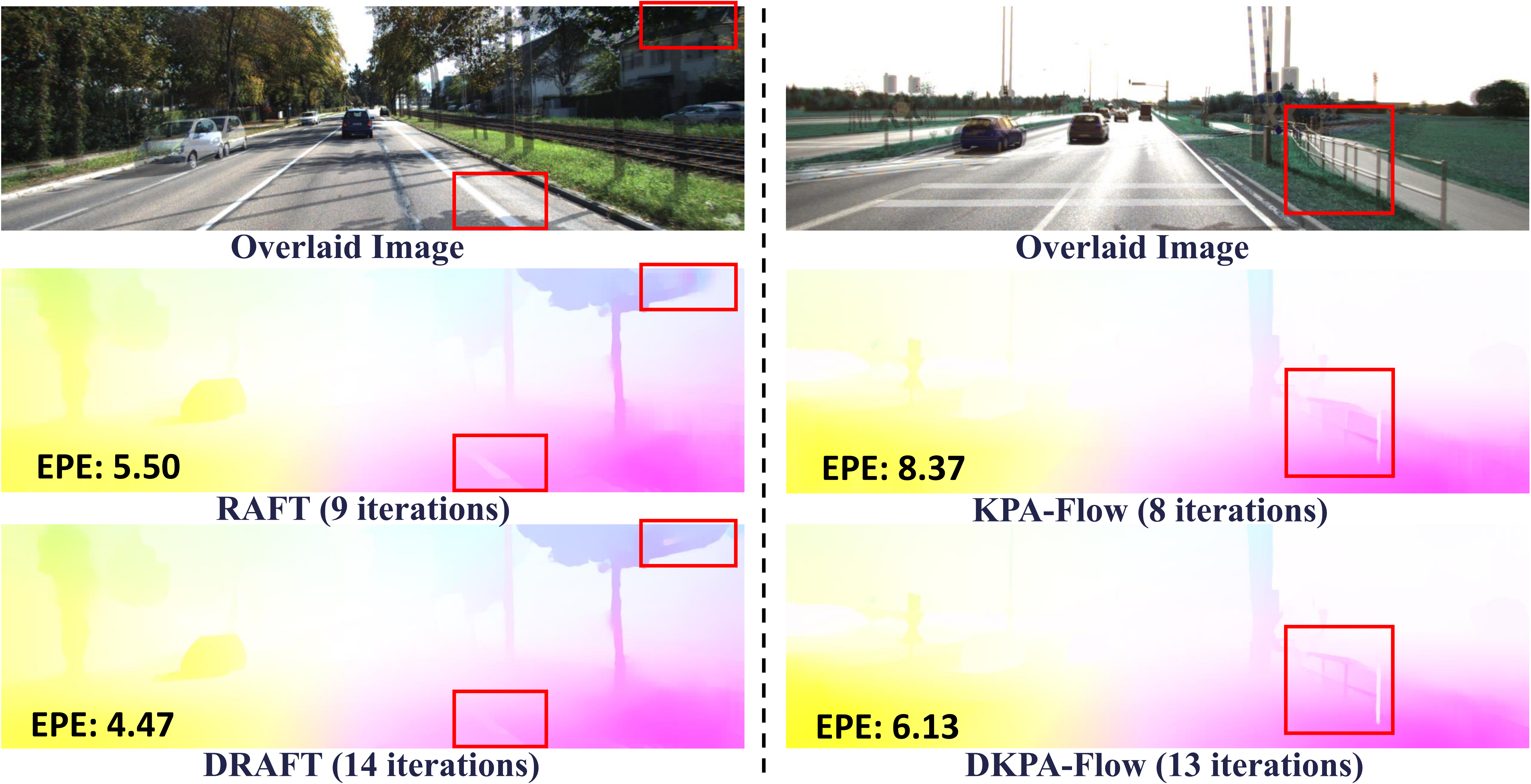}
    \caption{Qualitative comparison on the KITTI-train dataset.
    }
    \label{009_compare_s}
    \end{figure}

    \begin{figure}[t]
    \centering
    \includegraphics[width=0.9\linewidth]{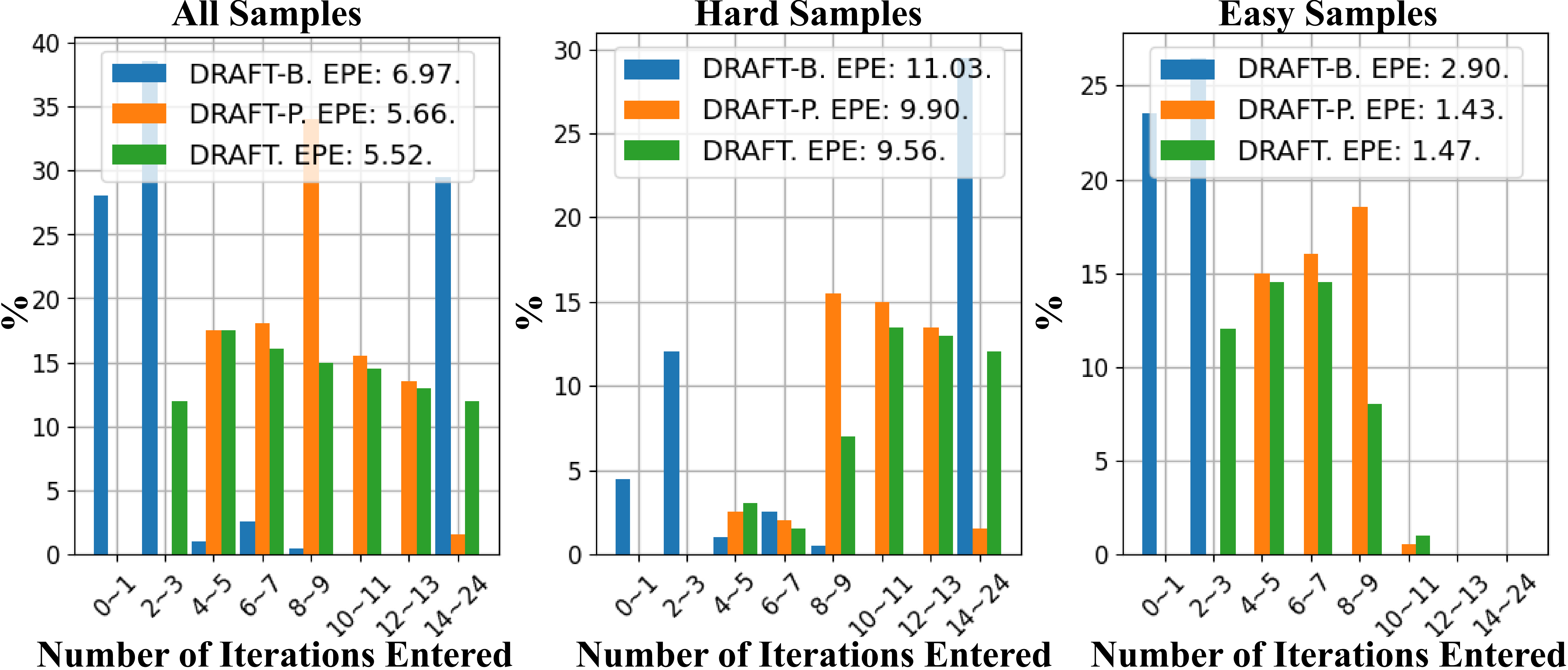}
    \caption{Ablation study of the contextual information on KITTI. We count the percentage of the total number of iterations for all samples, hard samples, and easy samples. The FLOPs of all samples for DRAFT-B, DRAFT-P, and DRAFT are 277.28G, 276.70G, and 276.26G.
    }
    \label{007_ablation_s}
    \end{figure}

\begin{table}[t]
    \centering
    \resizebox{\linewidth}{!}
{
    \begin{tabular}{lccccccc}
    \toprule
     \multirow{2}{*}{Method} & \multicolumn{2}{c}{Sintel-Clean (train)} &  \multicolumn{2}{c}{Sintel-Final (train)} & \multicolumn{3}{c}{KITTI-15 (train)}  \\
    \cmidrule(r){2-3} \cmidrule(r){4-5} \cmidrule(r){6-8} & EPE & FLOPs(G) & EPE & FLOPs(G) &  EPE & F1-all & FLOPs(G)  \\
    \hline
    \hline
    \multicolumn{8}{c}{C+T Training Data} \\
    \hline
    DRAFT & 1.48 &  406 & 2.67 & 502  & 5.06 & 17.5 & 473 \\
    DRAFT-together & 1.43  & 378  & 2.70 & 486 & 4.97 & 17.5 & 394 \\
    DKPA & 1.22 &  411 & 2.48 & 416  & 4.25 & 15.7 & 552 \\
    DKPA-together & 1.27 & 406 & 2.50 & 417 & 4.25 & 15.7 & 536 \\
    \hline
\toprule
    \end{tabular}
}
    \caption{Results of joint training from scratch.
}
    \label{table_together}
\end{table}

\section{Overall Algorithm}
We provide the algorithm for the training phase in Algorithm~\ref{alg:dofn_train}.
The encoder first extracts the feature $\phi_{0}$, and then the 4D cost volume $C$ is constructed from the feature $\phi_{0}$.
Then the flow is updated $T$ times using the update operator.
Specifically, we input the feature $\phi_{t-1}$, flow field $f_{t-1}$, and 4D cost volume $C$ into the update operator to obtain refined feature $\phi_{t}$ and flow $f_{t}$, where $t \in\{1,2,...,T \}$. The flow field $f_0$ and history hidden cell $h_0$ are both initialized to 0 everywhere, and $\hat{\phi}_1$ and $\hat{f}_1$ are set to $\phi_{1}$ and $f_{1}$, respectively.
After executing the update operator, the feature $\hat{\phi}_1$ and $f_t$ are obtained using $p_{t-1}$ to aggregating the $\phi_{t}$, $\hat{\phi}_{t-1}$ and $f_{t}$, $\hat{f}_{t-1}$, respectively.
$p_{t}$ is generated by our proposed context-aware iteration policy network, and we input $\hat{\phi}_{t}, h_{t-1}, e_t, r$ into the policy network to get $h_t, p_t, i_t$. 
$r$ is a resource preference value sampled at random between 0.2 and 1.0.
We perform the policy network after each execution of the update network except for the $T$-th execution.

In addition, we also provide the algorithm for inference phase of our dynamic optical flow network in Algorithm~\ref{alg:dofn_test}. Instead of randomly sampling $r$ between 0.2 and 1.0 as in the training phase, we set $r$ based on the computational resource. In addition, we enter the update operator if $P_0 \textgreater P_1$ unless we skip the update operator.

\section{More Comparison Experiments}

To further understand how our dynamic network operates, we compare our DRAFT and DKPA-Flow models to RAFT \cite{10.1007/978-3-030-58536-5_24} and KPA-Flow \cite{Luo_2022_CVPR} with comparable FLOPs in the Figure~\ref{011_compare_s}.
We rank the difficulty of estimating optical flow for samples based on their EPE values using RAFT or KPA-Flow. The more difficult it is to estimate the flow, the larger EPE is, and vice versa.
Then, we divide samples evenly into the easy and hard categories.
Next, we calculate the percentage of the iterations for the easy and hard sample.
We find that our policy network assigns more iterations to estimate the flow of hard samples but fewer iterations to easy samples.
DRAFT and DKPA-Flow have 0.09 and 0.14 more EPE than RAFT and KPA-Flow for easy samples but have 0.66 and 0.57 less EPE for hard samples.
In conclusion, our policy network assigns more iterations to samples whose flow is significant improved in subsequent iterations.
Consequently, DRAFT and DKPA-Flow outperform RAFT and KPA-Flow on average across all (easy and hard) samples.
In addition, we display the visualization results in Figure~\ref{009_compare_s}, where we can observe that assigning more iterations to hard samples improves optical flow significantly.

\section{More Ablation Studies}
We display the FLOPs of each ablation model in Table~\ref{table:ablation_s}. We can see that our DRAFT or DFlowformer outperforms other ablation models and achieves the lowest FLOPs simultaneously.

\noindent
\textbf{Ablation Study of Training Together.}
The policy network can be trained together with the original model, and we provide these results in Table~\ref{table_together}. It shows that training together leads to the same or better performance.

\noindent
\textbf{Ablation Study of Skipping Strategy.}
The skipping strategy introduced in the origin paper is to determine whether to skip the iteration or not at each timestep.
A particular skipping strategy is to end the optical flow estimation after one skip iteration. 
Therefore, we build an ablation model denoted as `-Exit' using this method. Table~\ref{table:ablation_s} shows that our DRAFT and DFlowFormer perform better than DRAFT-Exit. For example, DFlowFormer achieves 0.14 and 1.1 reductions for EPE and F1-all values on KITTI compared to DRAFT-Exit.
This might be because `-Exit' is probably making exit decisions wrongly at the early timestep.

In our ablation study, we also employ alternative methods to determine the optimal number of iterations for each sample. Using an early exiting strategy, we build an ablation model denoted as `-Exit.' Early exiting indicates that the network determines the earliest to exit the iteration permanently.
As shown in Table~\ref{table:ablation_control}, our DRAFT and DFlowFormer perform better than DRAFT-Exit. For example, DFlowFormer achieves 0.14 and 1.1 reduced for EPE and F1-all values on KITTI compared to DRAFT-Exit.
This might be because our re-enter strategy can prevent wrong exit decisions.

\noindent
\textbf{Ablation Study of Contextual Information.}
To further understand how contextual information works, we calculated the percentage of iterations entered for all samples, hard samples and easy samples, respectively. As shown in Figure~\ref{007_ablation_s}, we can observe that the optimal number of iterations predicted by the `-B' model is either extremely small or extremely large, indicating that the network is incapable of making accurate predictions in the absence of contextual information. Furthermore, we find that our DRAFT assigns more iterations to hard samples and reduces EPE by 0.34 while assigning fewer iterations to easy samples and increasing EPE by 0.04.
This is beneficial for our dynamic network since, on average, hard samples reduce more EPE than easy samples.

\end{document}